\documentclass[conference]{IEEEtran}
\usepackage{times}
\pdfoutput=1
\usepackage{multicol}
\usepackage{amsfonts}
\usepackage{amsmath} 
\usepackage{subcaption}
\usepackage{graphicx}
\usepackage{multirow}
\usepackage{booktabs}
\usepackage[font=small]{caption}
\usepackage[linesnumbered, commentsnumbered, ruled, lined, boxed]{algorithm2e}
\usepackage[dvipsnames]{xcolor}
\usepackage[backref,pageanchor=true,plainpages=false, pdfpagelabels, bookmarks,bookmarksnumbered, 
pdfborder=0 0 0,  
]{hyperref}
\hypersetup
{
  colorlinks=true,
  linkcolor=blue!60!black,
  urlcolor=blue!60!black,
  citecolor=blue!60!black
}
 
\newcommand{\argmin}{\operatornamewithlimits{arg\,min}}

\newcommand{\eg}{\textit{e.g.}~}

\newcommand{\figref}[1]{Fig.~\ref{#1}}
\newcommand{\tableref}[1]{Table ~\ref{#1}}
\newcommand{\sref}[1]{Section~\ref{#1}}
\newcommand{\eqnref}[1]{Eq.~\ref{#1}}
\newcommand{\algref}[1]{Alg.~\ref{#1}}

\newcommand{\bel}{\text{bel}}
\newcommand{\state}{\mathbf{q}}
\newcommand{\ballcenter}{\mathbf{p}}
\newcommand{\State}{Q}
\newcommand{\action}{\mathbf{u}}
\newcommand{\Action}{U}
\newcommand{\obs}{\mathbf{z}}
\newcommand{\Obs}{Z}
\newcommand{\obsc}{\mathbf{c}}
\newcommand{\obsq}{{\mathbf{q}_\text{e}}}
\newcommand{\offset}{{\Delta \state}}
\newcommand{\weight}{w} 
\newcommand{\particles}{Q}

\newcommand{\Obstacles}{X_\text{obs}}
\newcommand{\Boundary}{X_\text{surf}}
\newcommand{\Physics}{\textbf{F}}

\DeclareMathOperator{\distance}{dist}
\DeclareMathOperator{\interior}{int}
\DeclareMathOperator{\project}{proj}
\DeclareMathOperator{\uniform}{uniform}
\DeclareMathOperator{\Prob}{Pr} 
\IEEEoverridecommandlockouts
\begin{document}

\title{The Manifold Particle Filter for State Estimation on High-dimensional Implicit Manifolds}

\author{
\IEEEauthorblockN{Matthew Klingensmith} 
\thanks{All authors are with the Carnegie Mellon Robotics Institute, 5000 Forbes Avenue, Pittsburgh PA}
		\IEEEauthorblockA{{mklingen@andrew.cmu.edu}}
\and
\IEEEauthorblockN{Michael C. Koval}
	\thanks{This work was supported by a NASA Space Technology Research Fellowship (award NNX13AL62H), the National Science Foundation (award IIS-1218182  and IIS-1409003), the U.S Office of Naval Research grant No. N000141210613 and NSF grant No. IIS-1426703 and the Toyota Motor Corporation.}
	\IEEEauthorblockA{{ mkoval@cs.cmu.edu}}
\and
\IEEEauthorblockN{Siddhartha S. Srinivasa}
	\IEEEauthorblockA{{siddh@cs.cmu.edu}}
\and
\IEEEauthorblockN{Nancy S. Pollard}
	\IEEEauthorblockA{{nsp@cs.cmu.edu}}
\and
\IEEEauthorblockN{Michael Kaess}
	\IEEEauthorblockA{{kaess@cmu.edu}}
}



%

\maketitle

\begin{abstract}
We estimate the state a noisy robot arm and underactuated hand using an Implicit Manifold Particle Filter (MPF) informed by touch sensors. As the robot touches the world, its state space collapses to a contact manifold that we represent implicitly using a signed distance field. This allows us to extend the MPF to higher (six or more) dimensional state spaces. Earlier work (which explicitly represents the contact manifold) only shows the MPF in two or three dimensions \cite{koval2015manifold_ijrr}. Through a series of experiments, we show that the implicit MPF converges faster and is more accurate than a conventional particle filter during periods of persistent contact. We present three methods of sampling the implicit contact manifold, and compare them in experiments.
\end{abstract}

\IEEEpeerreviewmaketitle

\section{Introduction} 

\begin{figure} %
  \centering%
  \begin{subfigure}{0.5\columnwidth}%
    \centering%
    \includegraphics[width=\textwidth]{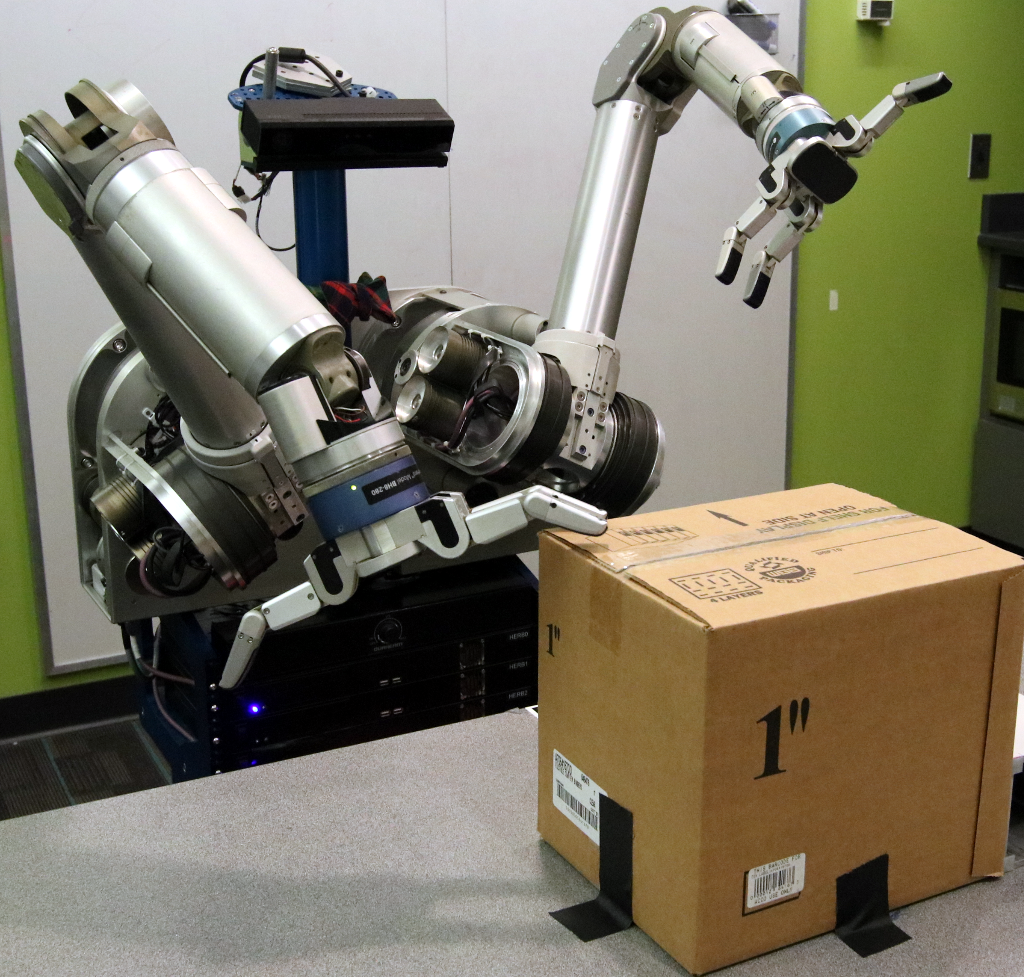}%
  \end{subfigure}%
  \begin{subfigure}{0.5\columnwidth}%
    \centering%
    \includegraphics[width=\textwidth]{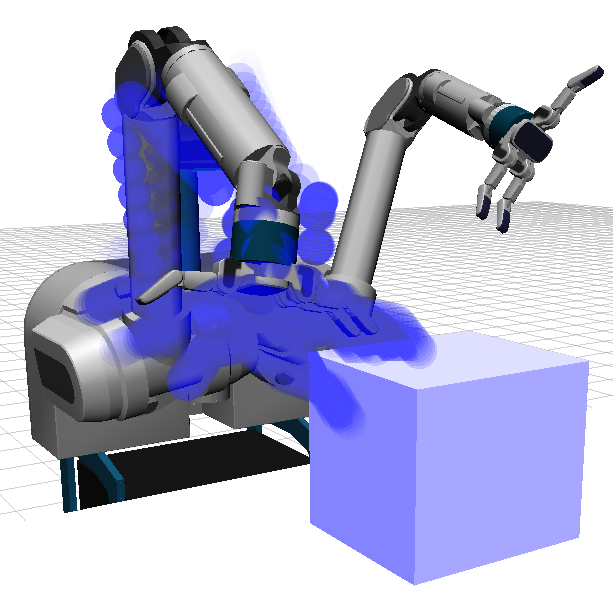}%
  \end{subfigure}%
  \\
  \begin{subfigure}{0.5\columnwidth}%
    \centering%
    \includegraphics[width=\textwidth]{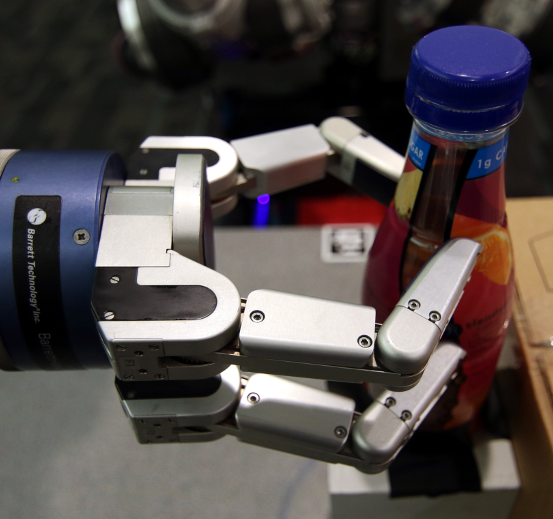}%
  \end{subfigure}%
  \begin{subfigure}{0.5\columnwidth}%
    \centering%
    \includegraphics[width=\textwidth]{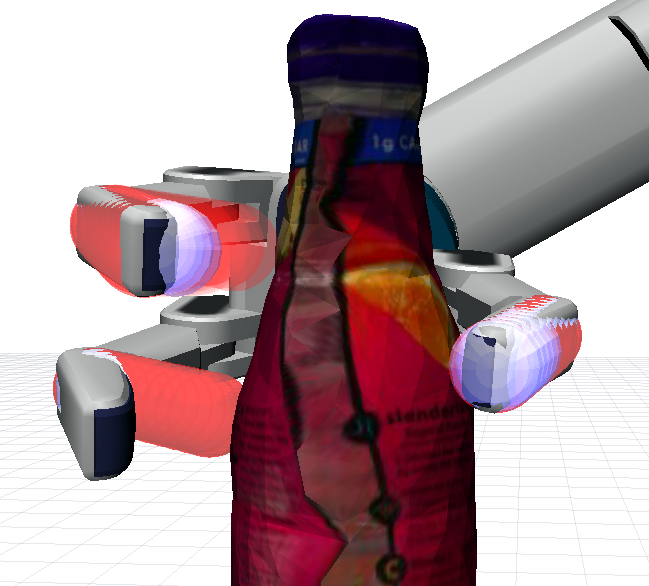}%
  \end{subfigure}%
  \caption{%
    Two examples of imperfect proprioception.
    (Top) The Barrett WAM~\cite{salisbury1988preliminary} touching a box. The
    solid render shows the configuration of the arm estimated by the WAM's
    encoders and the blue renders are particles. Note that there are two modes
    in the distribution.
    (Bottom) The BarrettHand~\cite{townsend2000barretthand} grapsing a bottle
    with no position sensors on its distal joints.  The solid render shows the
    configuration of the hand when the distal joints are assumed to be fixed.
    The red render shows the mode estimated from the particles.
  }%
  \label{fig:robot_experiments}%
\end{figure}

Robots often have imperfect proprioception. This may arise from
difficult-to-model transmissions, underactuated degrees of freedom, or poorly
calibrated sensors. \figref{fig:robot_experiments}~(Top) shows a Barrett WAM
arm~\cite{salisbury1988preliminary} touching a box. The WAM measures its joint
angles through a cable drive transmission that suffers from hysteresis related
to stretch in the cables~\cite{boots2014learning}.
\figref{fig:robot_experiments}~(Bottom) shows the
BarrettHand~\cite{townsend2000barretthand} grasping a bottle.  The position of
the distal finger joints depends on the state of a mechanical clutch that
engages when a torque threshold is met.

In both cases, the nominal configuration reported by the robot is in error. In
\figref{fig:robot_experiments}~(Top), the robot believes that it is several
centimeters above the box even though a finger is in contact. In
\figref{fig:robot_experiments}~(Bottom), the robot computed the configuration of
its distal links under the assumption that the clutch did not engage even
though all three fingers made contact with the object. In both cases, the
nominal joint positions are inconsistent with the robot's readings from its
contact sensors.

\emph{Our goal is to use contact sensors to refine a robot's estimate of its
configuration.} Like related work (\sref{sec:related_work}) dating back
to the 1970s~\cite{simunovic79information}, we frame this as a problem of
Bayesian estimation. State is the configuration of the robot, an action is a
commanded change in configuration, and an observation is a measurement from the
robot's contact sensors (\sref{sec:background}). Under this formulation, a
contact observation constrains the set of feasible states to a lower
dimensional \emph{contact manifold} that place the active sensors in
non-penetrating contact with the environment.

Traditional Bayesian estimation techniques perform poorly on this domain. The
extended~\cite{kalman1960kf} and unscented~\cite{julier1997ukf} Kalman filters
assume a Gaussian distribution over state, which cannot accurately represent a
set of feasible solutions on the contact manifold. The conventional particle
filter (CPF)~\cite{gordon1993novel} suffers from particle deprivation because
there the contact manifold has zero measure: there is zero probability of
sampling a state from the ambient space that lies on
it~\cite{koval2015manifold_ijrr}.

Instead, we use the \emph{manifold particle filter}
(MPF)~\cite{koval2015manifold_ijrr}. The MPF is identical to the CPF when the
robot has not sensed contact. When the robot does sense contact, the MPF
samples particles from the contact manifold and re-weights them based on their
proximity to the previous set of particles. The MPF has been successfully used
to estimate the pose of an object relative to the hand during planar
manipulation using explicit analytic and sample-based representations of the
contact manifold.

 
Explicit representations of the contact manifold do not scale to the
high-dimensional state space of this problem. Our key insight is to build an
\emph{implicit representation of the contact manifold} using a signed distance
field (SDF) and constraint projection (\sref{sec:implicit_mpf}). First,
we sample a set of particles uniformly from the region of state space near the
previous set of particles. Then, we use a SDF and the Jacobian of the
manipulator to project the samples onto the manifold.

We demonstrate the efficacy of this technique in simulation
(\sref{sec:experiments}) and real robot (\sref{sec:robot_experiments})
experiments for two problem domains. First, we consider a two, three, or seven
degree-of-freedom arm with noisy positions sensors moving in a static
environment. Second, we consider an underactuated robotic hand grasping a
static object. In both cases, we show that the proposed technique significantly
outperforms the CPF.

We believe that the proposed approach is applicable to a wide variety of
problem domains. However, it has one key limitation: it requires a known,
static environment. We plan to relax this assumption in future work by
incorporating the pose of dynamic objects into the filter's state space
(\sref{sec:limitations}).

\section{Related Work}
\label{sec:related_work}

There are a variety of approaches that use feedback from contact sensors for
manipulation. One approach is to plan a sequence of move-until-touch actions
that are guaranteed to localize an object to the desired
accuracy~\cite{petrovskaya2011global,javdani2013efficient,hebert2013next}
Other approaches formulate the problem as a partially observable Markov
decision process~\cite{smallwood1973optimal} and solve for a policy that
optimizes expected
reward~\cite{hsiao2007grasping,horowitz2013interactive,koval2014precontact_rss}.
These algorithms require an efficient implementation of a Bayesian state
estimator that can be queried many times during planning. Our approach could be
used as a state estimator in one of these planners.

Recent work has used the conventional particle filter
(CPF)~\cite{gordon1993novel} to estimate the pose of an object while it is
being pushed using visual and tactile
observations~\cite{zhang2012application,zhang2013dynamic}. Unfortunately, the
CPF performs poorly because the set of feasible configurations lie on
lower-dimensional contact manifold. The manifold particle filter (MPF) avoids
particle deprivation by sampling from an approximate representation of this
manifold~\cite{koval2015manifold_ijrr}. However, building an explicit
representation of the contact manifold is only feasible for
low-dimensional---typically planar---problems.

In this work, we extend the MPF to estimate the full---typically six or more
dimensional---configuration of a robot under proprioceptive uncertainty.
Depth-based trackers, such as articulated
ICP~\cite{pellegrini2008generalisation,krainin2011manipulator},
GMAT~\cite{klingensmith2013closed}, and DART~\cite{schmidt2014dart}, can track
the configuration of a robot using commercially available depth sensors. DART
has been extended to incorporate contact observations using a method similar to
our constraint projection \cite{schmidt2015depth}. However, all of these
methods maintain a uni-modal state estimate and, thus, perform best when the
robot is visible and un-occluded. In contrast, the MPF maintains a full
distribution over belief space, only requires contact sensors, and is
unaffected by occlusion.

The work most similar to our own aims to localize a mobile robot in a known
environment using contact with the environment. Prior work has used the CPF to
localize the pose of a mobile manipulator by observing where its arm contacts
with the environment~\cite{dogar2010proprioceptive}. The same approach was used
to localize a quadruped on known terrain by estimating the stability of the
robot given its configuration~\cite{chitta2007proprioceptive}. Estimating the
base pose of a mobile robot is equivalent to solving the problem formulated in
this paper with the addition of a single, unconstrained six degree-of-freedom
joint that attaches the robot to the world.

Techniques for estimating the configuration of an articulated body from contact
sensors are applicable to humans as well as robots. Researchers in the computer
graphics community have instrumented objects with contact
sensors~\cite{kry2006interaction} and used multi-touch
displays~\cite{chung2015quadratic} to reconstruct the configuration of a human
hand from contact observations. Both of these approaches generate natural hand
configurations by interpolating between data points collected on the same
device. Other work has used machine learning to reconstruct the configuration
of a human from the ground reaction forces measured by pressure
sensors~\cite{ha2011human}.  We hope that our approach is also useful in these
problem domains.

A recent work on robot state estimation from touch~\cite{roncone2014automatic} 
estimates the robot's joint angles and Denavit Hartenberg parameters from a single 
self-touch. Our work is related only insofar as we are estimating the joint angles
of a robot arm using touch, but differs in that we use online filtering rather than
batch optimization, and rather than estimating the state using one closed-chain self
touch, we estimate the state using multiple open-chain touches of the environment.

\section{Background}
\label{sec:background}

Our approach is straightforward: as the robot moves around and contacts surfaces, 
we run a high-dimensional Manifold Particle Filter in its configuration space. Physical 
constraints from contact and collision with the robot's body allow us to reason 
about how its joints have moved, compensating for joint angle noise.

\subsection{Problem Definition}
\label{sec:problem_definition}

Consider a robot with configuration space $\State = \mathbb{R}^n$. At each
timestep the robot executes a control input $\action \in \Action$, transitions
to the successor state $\state' \sim p(\state' | \state, \action)$, and
receives an observation $\obs \sim p(\obs | \state', \action)$. An observation
$\obs = (\obsq, \obsc) \in \Obs$ includes a noisy estimate $\obsq \in
\mathbb{R}^n$ of the robot's configuration and a binary vector of readings
$\obsc \in \{ 0, 1 \}^m$ from the robot's $m$ contact sensors.

Our goal is to estimate the \emph{belief state} $\bel(\state_t) = p(\state_t |
\action_{1:t}, \obs_{1:t})$, the probability distribution over the state
$\state_t$ given the history of actions $\action_{1:t} = \action_1, \dotsc,
\action_t$ and observations $\obs_{1:t} = \obs_1, \dotsc, \obs_t$.

\subsubsection{Transition Model}
Our method is applicable to any transition model where it is possible to sample
from $p(\state' | \state, \action)$ for given values of $\state$ and $\action$.
In our experiments, we choose $\action$ to be commanded joint velocities and
define
\begin{align*}
  \state' = \Physics\left(\state + \uniform B\left(\action, r_a\right) \Delta t\right)
\end{align*}
as the noisy forward integration of the control input $\action$, where the noise
is a uniform sample drawn from a ball $B$ of radius $r_a$. We assume that 
the world is static and does not change in response to the robot's touch. In our
experiments we use a simple physics simulation $\Physics$ involving frictionless soft collisions
with the environment. As the robot touches the static environment, contact forces
push it away from obstacles.

\subsubsection{Observation Model}
We assume that proprioceptive and contact sensor observations are conditionally
independent given the state $\state$ and most recent action $\action$. Under
this assumption, we can express the observation model
\begin{align*}
  p(\obs | \state, \action) &= p(\obsq | \state, \action) p(\obsc | \state, \action)
\end{align*}
as the product of two marginal distributions.

The distribution $p(\obsq | \state, \action)$ models uncertainty in the robot's joint
position sensors. We make no assumptions about the form of this distribution
other than that it be possible to evaluate the probability density for given
values of $\obsq$, $\state$, and $\action$. In our experiments, we model $\obsq$ as a
measurement of $\state$ corrupted by a static (but unknown) joint offset 
\begin{align*}
\state = \obsq + \offset
\end{align*}
the initial offset $\offset$ is sampled from a Gaussian at time $0$:
\begin{align*}
	\offset \sim \mathcal{N}(0, \Sigma_\offset)
\end{align*}
and remains static for the rest of the experiment. Therefore
\begin{align*}
p(\obsq | \state, \action) = \mathcal{N}(\obsq - \state, \Sigma_\offset)
\end{align*}
To model this, instead of estimating a belief over the full state $\state$, we estimate 
a belief over offsets $\offset$ (the state being derivative of the offset and the joint encoders).
It should be understood that whenever the state $\state$ is mentioned in this work, what is meant
is $\state = \obsq + \offset$. 

If one or more joints are unobserved, such as the distal joints in
\figref{fig:robot_experiments}~(Bottom), then $\obsq$ has fewer dimensions
than $\state$. We treat those unobserved dimensions of $\state$ as initially having 
uniform probability over the entire state space.

The distribution $p(\obsc | \state, \action)$ models the robot's contact sensors. Each
sensor is a rigid body that is attached to one of the robot's links. The sensor
returns ``contact'' if any part of the sensor touches the environment and
otherwise returns ``no contact.'' Similar to prior
work~\cite{koval2015manifold_ijrr}, we assume that the contact sensors do not
generate false positives.

\subsection{Bayes Filter}
The \emph{Bayes filter} provides method of recursively constructing
$\bel(\state_t)$ from $\bel(\state_{t-1})$. Given an initial belief $\bel(\state_0)$, the
Bayes filter applies the update rule
\begin{align*}
  \bel(\state_t) &= \eta \, p(\obs | \state_t, \action)
    \int_Q p(\state_t | \state_{t-1}, \action) \bel(\state_{t-1})
    \, d \state_{t-1}
\end{align*}
where $\eta$ is a normalization constant. This equation is derived from the
definition of the belief state and the Markov property.

There are several ways of implementing the Bayes filter. The discrete Bayes
filter represents $\bel(\state_t)$ as a piecewise constant histogram.
Discretization is intractable on our problem $\State$ is typically high
dimensional: $n \ge 6$ for a most manipulators. The Kalman filter, extended
Kalman filter~\cite{kalman1960kf}, and unscented Kalman
filter~\cite{julier1997ukf} avoid discretization by assuming that
$\bel(\state_t)$ is Gaussian. This assumption is not valid for our problem: the
observation model $p(\obsc | \state, \action)$ is discontinuous and tends to
produce multi-modal belief states.

\subsection{Conventional Particle Filter}
Instead, we use the particle filter~\cite{gordon1993novel}. The \emph{particle
filter} (\algref{alg:cpf}) is an implementation of the Bayes filter that
represents $\bel(s_t)$ using a discrete set of weighted samples $\particles_t =
\{ \langle \state_t^{[i]}, \weight_t^{[i]} \rangle \}_{i = 1}^k$, known as
\emph{particles}. The set of particles $Q_t$ at time $t$ is recursively
constructed from the set of particles $X_{t-1}$ at time $t - 1$ using
importance sampling.

\begin{algorithm}[t]
  \KwIn{$\particles_{t-1}$ particles sampled from $\bel(\state_{t-1})$}
  \KwOut{$\particles_{t}$ particles sampled from $\bel(\state_t)$}

  $Q_t \gets \emptyset$ \\
  \For{$\state_{t-1}^{[i]} \in \particles_{t - 1}$}{
    $\state_t^{[i]} \sim p(\state_t | \state_{t-1}^{[i]}, u_t)$ \\
    $w_t^{[i]} \gets p(\obs | \state_t^{[i]}, u_t)$
  }
  $\particles_t \gets \textsc{Resample}(\particles_t)$
  \caption{\sc Conventional Particle Filter}
  \label{alg:cpf}
\end{algorithm}

First, the particle filter samples a set of $k$ states $\state_t^{[i]} \sim
\rho_\text{conv}(\state)$ from a \emph{proposal distribution}
$\rho_\text{conv}(\state)$. Conventionally, the proposal distribution is chosen
to be 
\begin{align}
  \rho_\text{conv}(\state_t) &= \int_Q p(\state_t | \state_{t-1}, \action)
    \bel(\state_{t-1}) \, d \state_{t-1},
  \label{eqn:proposal_conventional}
\end{align}
the transition model applied to the previous belief state. This can be
implemented by forward simulating $\particles_{t-1}$ to time $t$ using the
transition model.

Next, the particle filter computes an \emph{importance weight} $\weight_t^{[i]}
= \bel(\state_t^{[i]}) / \rho_\text{conv}(\state_t^{[i]})$ to correct for the
discrepancy between the proposal distribution $\rho_\text{conv}(\state_t)$ and
the desired distribution $\bel(\state_t)$.  When using the proposal
distribution shown in \eqnref{eqn:proposal_conventional}, the corresponding
importance weight s $\weight_t^{[i]} = \eta \, p(\obs_t | \state_t^{[i]},
\action_t)$. This can be thought of as updating $\particles_t$ to agree with
the most recent observation $\obs_t$.

Finally, the particle filter periodically resamples each particle in
$\particles_t$ with replacement, with probability proportional to its weight.
This process is known as \emph{sequential importance resampling} (SIR) and is
necessary to achieve good performance over long time horizons.

\subsection{Degeneracy of the Conventional Particle Filter}
\label{sec:background-degeneracy}
Prior work~\cite{koval2015manifold_ijrr} has shown that the conventional
particle filter (CPF) performs poorly with contact sensors because $\bel(s_t)$
collapses to a lower-dimensional manifold. This leads to \emph{particle
deprivation} during contact, where $\weight_t^{[i]} = 0$ for all but a few
particles, because it is vanishingly unlikely that a particle sampled from the
transition model will lie on the zero measure contact manifold.

To see why this is the case, consider a 2D, two jointed and two linked robot
with a single point contact sensor on its distal link \figref{fig:manifolds}.
When the robot contacts the environment, the contact state of its sensor changes.
Infinitesimal motion along the surface results in the same contact state, but
Infinitesimal motion away from the surface results in a different contact state.
The set of configurations with the same contact state locally form a manifold that
is lower dimensional than the full state space.

\subsection{Manifold Particle Filter}
The \emph{manifold particle filter} (MPF, \algref{alg:mpf}) avoids particle
deprivation by operating in two modes~\cite{koval2015manifold_ijrr}. When no
contact is observed, particle deprivation is behaves identically to the CPF by
sampling particles from the transition model and weighting them by the
observation model. When contact is observed, the MPF switches to sampling
particles from the observation model and weighting them by the transition
model.

Both modes of the MPF implement importance sampling with different proposal
distributions. During contact, the MPF samples particles from the \emph{dual
proposal distribution}
\begin{align*}
  \rho_\text{dual}(\state_t)
    &= \frac{p(\obs_t | \state_t, \action_t)}{p(\obs_t | \action_t)},
\end{align*}
where $p(\obs_t | \action_t)$ is a normalization constant. Sampling from
$\rho_\text{dual}(\state_t)$ generates configurations that are consistent with
the most recent observation $\obs_t$ according to the observation model. The
remainder of this paper describes how to generate these samples efficiently. 

The importance weight for a particle $\state_t^{[i]} \sim
\rho_\text{dual}(\state_t)$ is
\begin{align}
  w_t^{[i]} &= \eta \int_Q p(\state_t | \state_{t-1}, \action)
    \bel(\state_{t-1}) \, d \state_{t-1}
  \label{eqn:dual-weight}
\end{align}
where $\eta$ is another normalization constant. The importance weight
$w_t^{[i]}$ incorporates information from $\bel(\state_{t-1})$ into the
posterior belief state; i.e. enforces temporal consistency with the transition
model.

Computing $w_t^{[i]}$ exactly is not possible with a particle-based
representation of $\bel(\state_{t-1})$. Instead, we forward simulate the
particles $\particles_{t-1}$ by applying the transition model just like we do
in the CPF. This set of particles $\particles_{t-1}^+$ are distributed
according to the right-hand side of \eqnref{eqn:dual-weight}. We approximate
the weight $w_t^{[i]}$ using a kernel density
estimate~\cite{rosenblatt1956remarks} built from $\particles_{t-1}^+$.

\begin{algorithm}[t]
  \KwIn{$\action_t$, $\particles_{t-1}$ particles sampled from $\bel(\state_{t-1})$}
  \KwOut{$\particles_{t}$ particles sampled from $\bel(\state_t)$}

  $\particles_t \gets \emptyset$ \\
  \For{$\state_{t-1}^{[i]} \in \particles_{t-1}$}{
    \uIf{$\obsc_t = 0$}{
      $\state_t^{[i]} \sim p(\state_t | \state_{t-1}^{[i]}, \action_t)$ \\
      $w_t^{[i]} \gets p(\obsc_t | \state_t^{[i]}, \action_t)$
    }\Else{
      $\state_t^{[i]} \sim \uniform M(\obsc)$ \\
      $w_t^{[i]} \gets \textsc{KernelDensityEstimate}(
        \particles_{t-1}^+, \state_t^{[i]})$
    }
  }
  $\particles_t \gets \textsc{Resample}(\particles_t)$

  \caption{\sc Manifold Particle Filter}
  \label{alg:mpf}
\end{algorithm}

\section{Implicit Contact Manifold Representation}
\label{sec:implicit_mpf}
Implementing the MPF requires sampling from the lower-dimensional contact
manifold associated with the active contact sensors. In this section, we
formally define the contact manifold (\sref{sec:representation-manifold})
associated with contact observation $\obsc$ and explain why it is infeasible to
build an explicit representation of this manifold.

Instead, we implicitly define the contact manifold as the iso-contour of a loss
function (\sref{sec:representation-implicit}) and use a local optimizer to
project onto it (\sref{sec:representation-projection}) by using a
signed-distance field to compute gradient information
(\sref{sec:representation-sdf}).  Finally, we describe three different methods
of using projection to sample from the observation model
(\sref{sec:representation-sampling}).

\subsection{Contact Manifold}
\label{sec:representation-manifold}
Suppose the robot is in a static environment with obstacles $\Obstacles
\subseteq \mathbb{R}^3$ with boundary $\Boundary = \Obstacles \setminus
\interior(\Obstacles)$. The robot's $i$-th contact sensor is a rigid body with
geometry $c(\state) \subseteq \mathbb{R}^3$ in configuration $\state$. If the
robot senses contact with sensor $i$, then we know that $c_i(\state)$ is in
non-penetrating contact with the environment; i.e. $c_i(\state) \cap \Boundary
\ne \emptyset$.

\begin{figure}%
  \centering%
  \begin{subfigure}{\columnwidth}%
    \centering%
    \includegraphics[width=0.75\textwidth]{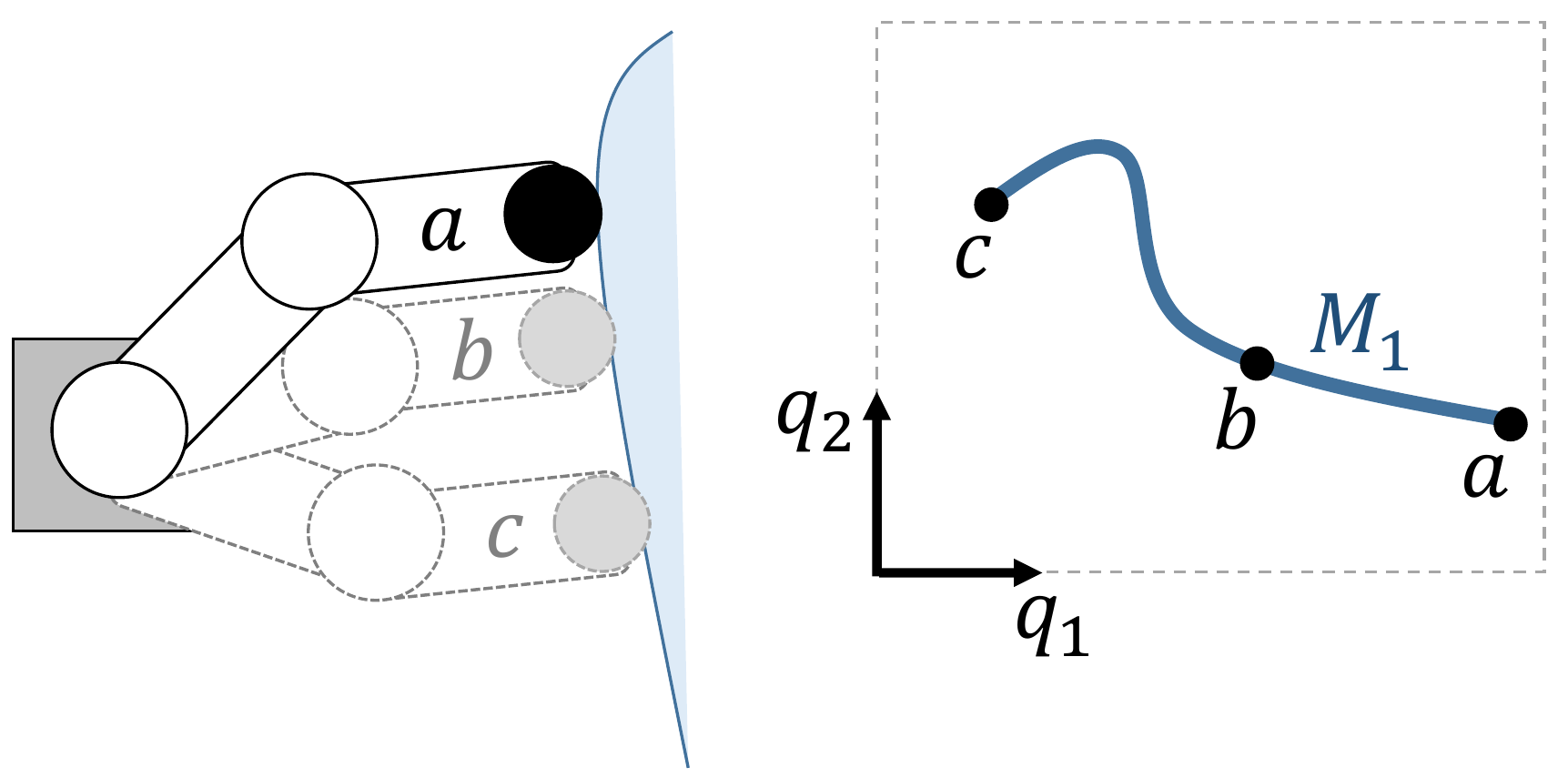}%
    \caption{Contact manifold}%
    \label{fig:manifold}%
  \end{subfigure}%
  \\
  \begin{subfigure}{\columnwidth}%
    \centering%
    \includegraphics[width=0.75\textwidth]{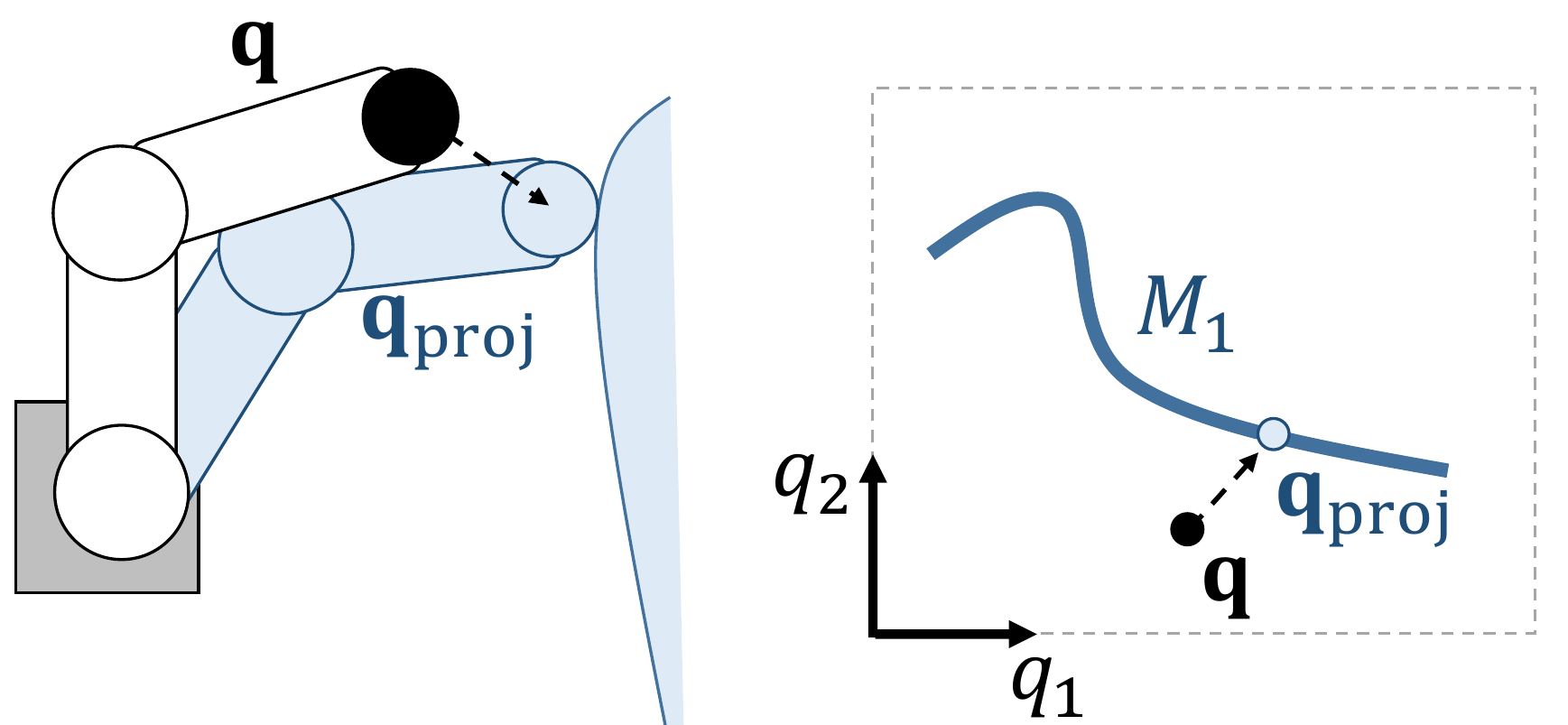}%
    \caption{Projection onto the contact manifold}%
    \label{fig:manifold_projection}%
  \end{subfigure}%
  \caption{%
    Example of the contact manifold for a contact sensor (black circle) on a
    planar, two degree-of-freedom robot. The robot's configuration space,
    parameterized by the two joint angles $q_1$ and $q_2$ is shown on the right.
    (\subref{fig:manifold})~Three different configurations $a$, $b$, and
    $c$ that lie on the contact manifold $M_1$.
    (\subref{fig:manifold_projection})~The configuration $\state$ is projected
    to a nearby configuration $\state_\text{proj}$ that lies on $M_1$.
  }%
  \label{fig:manifolds}%
\end{figure}

We define the \emph{sensor contact manifold} $M_i$ of sensor $i$ as
\begin{align*}
  M_i &= \{ \state \in \State : c(\state) \cap \Boundary \ne \emptyset \},
\end{align*}
the set of all configurations that put $c_i(\state)$ in contact with the
environment. If multiple contact sensors are active, then we know that all
active sensors are in non-penetrating contact with the environment.
\figref{fig:manifold} shows a simple example of this.

The \emph{observation contact manifold} $M(\obsc)$ is given by the intersection
of the active sensor contact manifolds
\begin{align*}
  M(\obsc) &= \bigcap_{i \in \Phi(\obsc)} M_i,
\end{align*}
where $\Phi(\obsc)$ denotes the indices of the sensors active in $\obsc$.

Explicitly representing $M(\obsc)$ for small problems. For example,
\figref{fig:simulation-2dof} shows a 2D robot with two joints in an environment consisting
of a single point. In this environment, the contact manifold can be computed easily
using analytic inverse kinematics. However, as the environment and dimensionality
of the problem increase in complexity (\figref{fig:simulation-3dof}, \figref{fig:simulation-7dof}),
deriving an explicit representation of the manifold becomes computationally infeasible, 
because it requires computing inverse kinematics solutions for \textit{every} surface 
point which would cause the contact $\obsc$. 

\subsection{Implicit Representation of the Contact Manifold}
\label{sec:representation-implicit}
Luckily for the MPF, we do not need to compute an explicit representation of
$M(\obsc)$: we only have to be able to draw samples from it. In this work, we
first sample from the full state space, and then \textit{project} onto the
$M(\obsc)$, which is only represented implicitly as the iso-contour of a loss
function. We then reject any sample that is not close enough to the manifold.

We represent the sensor contact manifold $M_i$ as the zero iso-contour $M_i =
\{ \state \in \State : \distance(c_i(\state), \Obstacles) = 0 \}$ of the
\emph{signed distance function}
\begin{align*}
  \distance(X, Y) &= \min_{x \in X} \begin{cases}
    \hphantom{-}\distance(x, Y) &: x \not\in Y \\
    -\distance(x, Y) &: \text{otherwise}
  \end{cases}
\end{align*}
between the sensor and the environment. A signed distance is a positive value
equal to the distance between two disjoint sets or negative value equal to the
deepest penetration between two intersecting sets. The signed distance
$\distance(c_i(\state), \Obstacles)$ is zero iff contact sensor $i$ is in
non-penetrating contact with the environment in configuration $\state$.

A configuration $\state$ lies on the observation contact manifold $M(\obsc)$ if
the signed distance $\distance(c_i(\state), \Obstacles) = 0$ for all sensors
active $i \in \Phi(\obsc)$ in observation $\obsc$. We represent this set as the
zero iso-contour $M(\obsc) = \{ \state \in \State : D_{\obsc}(\state) = 0 \}$
of the loss function
\begin{align*}
  D_\obsc(\state)
    &= \sum_{i \in \Phi(\obsc)} \left[
      \distance(c_i(\state), \Obstacles)
    \right]^2,
\end{align*}
which is zero iff $\state \in M(\obsc)$. Any function that satisfies this
property is sufficient. We choose sum-of-squared distances to simplify the
projection operator described below.

\subsection{Projecting onto the Contact Manifold}
\label{sec:representation-projection}
We project a sample from the ambient space $\state$ onto the the contact
manifold $M(\obsc)$ by solving the optimization problem
\begin{align}
  \project(\tilde{\state}, \obsc)
    &= \argmin_{\state \in N(\tilde{\state})} D_\obsc(\state)
  \label{eqn:projection}
\end{align}
in a neighborhood $N(\tilde{\state}) \subseteq \State$ around an initial
configuration $\tilde{\state} \in \State$. If the distance $D_\obsc(\state) =
0$ at the end of the optimization, then we have found a configuration $\state
\in M(\obsc)$. \figref{fig:manifold_projection} shows an example of the outcome
of this process.

We implement the minimization in \eqnref{eqn:projection} using simple gradient descent
optimization. The optimizer is initialized with $\state^{(0)} = \tilde{\state}$ and iteratively 
applies the update rule 
\begin{align*}
  \state^{(j+1)} &= \state^{(j)}
    - \lambda \nabla D_\obsc^T D_\obsc ( \state^{(j)} )
\end{align*}
until $\state^{(j)}$ has converged, where $\lambda$ is the learning rate. Performing this update requires computing
the gradient
\begin{align*}
  \nabla D_\obsc &= \nabla_\state \distance \obsc(\state)
    &= 2 \sum_{i \in \Phi(\obsc)}
      \nabla_{\state}\distance(c_i(\state), \Obstacles),
\end{align*}
which, in turn, requires computing the gradient of the signed distance
function. We describe how to efficiently compute $\distance(\cdot, \cdot)$ and
$\nabla_{\state} \distance(\cdot, \cdot)$ in the next section.

Note that this procedure may converge to a configuration where $D_\obsc(\state)
\ge \epsilon$ because (1) there is no solution in the neighborhood
$N(\tilde{\state})$ or (2) the optimizer reached a local minimum. In either
case, the projection fails and needs to be re-initialized with a different
$\tilde{\state} \in \State$.

This method of projecting onto the contact manifold using an implicit representation
is commonly used in computer graphics to quickly compute contacts for simulated 
collision resolution and inverse kinematics. The method described here for computing
an implicit contact manifold for an articulated body is essentially the same as one 
described in~\cite{Tian2016116}.


\subsection{Signed Distance Computation}
\label{sec:representation-sdf}
Evaluating $D_\obsc(\state)$ requires computing the distance
$\distance(c_i(\state), \Obstacles)$ between each contact sensor $c_i(\state)$
and the environment $\Obstacles$. Computing this distance metric is difficult
and, potentially computationally expensive, for arbitrary geometric shapes. To
avoid this, we approximate the static environment, which may contain arbitrary
geometry, with a voxel grid and each contact sensors with a collection of
geometric primitives (e.g. spheres, capsules, boxes).

\begin{figure}%
  \centering%
  \includegraphics[width=0.75\columnwidth]{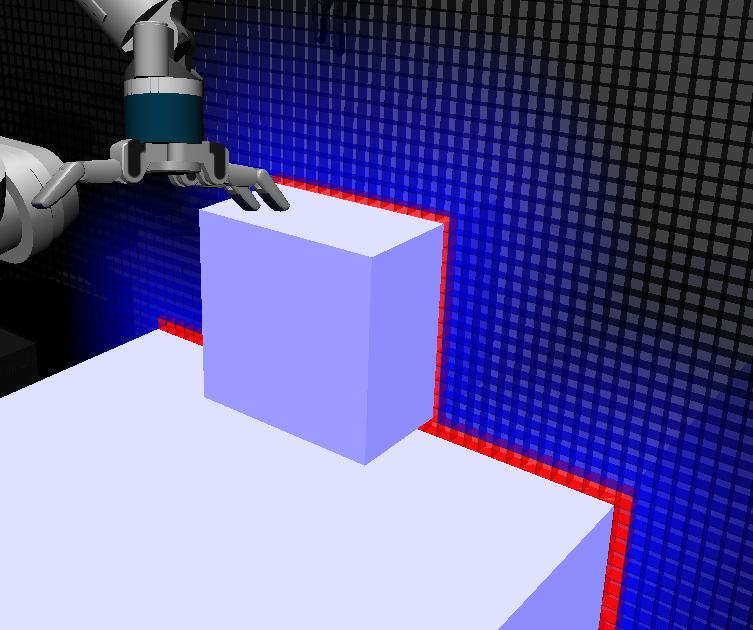}%
  \caption{%
    Visualizing a two-dimensional slice of the signed distance field (SDF)
    $\Phi(\cdot)$. Blue voxels have $\Phi(\cdot) > 0$ and red voxels have
    $\Phi(\cdot) \le 0$. Precomputing this SDF takes between two and five
    seconds. The voxel grid has a 2 cm resolution.
  }%
  \label{fig:sdf}%
\end{figure}

As an offline pre-computation step, we compute a discrete signed distance field
over the voxel grid using the technique from Felzenszwalb and
Huttenlocher~\cite{Felzenszwalb04}. A \emph{signed distance field} (SDF) is a
function $\Phi: x \in \mathbb{R}^3 \mapsto \distance(x, \Obstacles)$ that maps
each point $x$ in the workspace to its signed distance to the nearest obstacle.
\figref{fig:sdf} shows an example of a SDF computed in built in this way.

Without loss of generality, assume that the geometry of each contact sensor is
a sphere with center $\ballcenter_i(\state) \in \State$ and radius $r_i$. The
signed distance between the sensor and the environment is given by
\begin{align*}
  \distance(c_i(\state), \Obstacles)
    &= \Phi( \ballcenter_i(\state) ) - r_i.
\end{align*}
The gradient of the distance is given by
\begin{align*}
  \nabla_{\state} \distance(c_i(\state), \Obstacles)
  &= -\left[ \Phi(\ballcenter_i(\state)) - r_i \right]
     \left[ J_i(\state) \right]^T
     \nabla_{\mathbf{x}} \Phi(\ballcenter_i(\state)),
\end{align*}
where $J_i(\state) = \frac{\partial \ballcenter_i}{\partial \state}$ is the
linear Jacobian of the manipulator. We approximate the gradient
$\nabla_{\mathbf{x}} \Phi(\ballcenter_i(\state))$ of the SDF with a finite
difference.

Critically, evaluating $\Phi(\cdot)$ requires a single memory lookup and
evaluating the gradient $\Delta_x \Phi(x)$ can be efficiently approximated by a
finite difference. This is the same representation of the environment used by
CHOMP, a gradient-based trajectory optimizer~\cite{ratliff2009chomp}.

\subsection{Sampling from the Contact Manifold via Projection}
\label{sec:representation-sampling}
The projection operator described above starts with a single initial
configuration $\tilde{\state} \in \State$ and projects it onto the contact
manifold $M(\obsc)$. Sampling from the dual proposal distribution requires $n$
such samples distributed uniformly over $M(\obsc)$. We describe three different
approaches for selecting the set of initializations $\tilde{\State}_t = \{
\tilde{\state}^{[i]} \}_{i=1}^n$ used to generate the set of particles
$\State_t$. If a projection operation fails, i.e. $D_\obsc(\state^{[i]}) \ge
\epsilon$, then we generate a new initialization $\tilde{\state}^{[i]}$ and try
again.

\subsubsection{Uniform Projection}
The simplest strategy is to sample $\tilde{\state}^{[i]} \sim \uniform(\State)$
uniformly from the robot's configuration space. This method is unbiased with
respect to the previous set of particles $\particles_{t-1}$. Unfortunately,
since $\State$ is high dimensional, it may take a large number of particles to
adequately cover the the manifold. This may lead to particle deprivation.

\subsubsection{Particle Projection}
We can focus our samples near $\particles_{t-1}$ by directly projecting the
previous set of particles $\tilde{\particles}_t = \particles_{t-1}$ onto the
contact manifold. This method tightly focuses samples on the portions of the
manifold that will be assigned high importance weights. However, this comes
with two downsides: (1) $\particles_t$ will have a non-uniform distribution (2)
the set of particles may have size $|\particles_t| < n$ if projecting any
particles fails.
%

\subsubsection{Ball Projection}
We can combine the advantages of both approaches by uniformly sampling
particles $\tilde{\state}_t \sim \uniform R(\particles_{t-1})$ from the region
of configuration space $R(\particles_{t-1})$ near the previous set of particles
$\particles_{t-1}$. We define the region
\begin{align*}
  R(\particles_{t-1}) &= \bigcup_{\state^{[i]} \in \particles_{t-1}}
    B_{r_\epsilon}(\state^{[i]})
\end{align*}
where $B_{r_\epsilon}(\state^{[i]}) = \{ \state \in \State : ||\state -
\state^{[i]}|| < r_\epsilon \}$ is a ball centered at $\state^{[i]}$ with
radius $r_\epsilon$. The set $R(\particles_{t-1})$ is the union of all such
balls centered at the particles in $\particles_{t-1}$.
 
This approach is equivalent to particle projection as $r_\epsilon \to 0$.
and equivalent to uniform projection as $r_\epsilon \to \infty$. We select the
radius $r_\epsilon$ such that the transition model has probability
$\Prob(\state' \not\in R(\particles_{t-1})) < \epsilon$ of generating a
successor state $\state' \sim p(\state' | \state, \action)$ outside of
$R(\particles_{t-1})$ given any $\state \in \particles_{t-1}$. For this choice
of $r_\epsilon$, this approach is an $\epsilon$-approximation of the uniform
sampling method. However, this approach concentrates the particles
$\particles_t$ concentrated on the region of $M(\obsc)$ with high importance
weights.


\begin{figure*} 
  \centering
  \begin{subfigure}{2.3in}
    \centering 
    \includegraphics[width=1.0\textwidth]{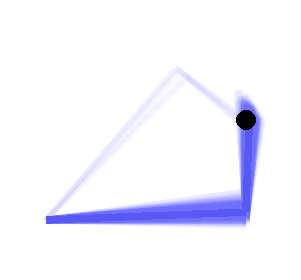} 
  \end{subfigure}
    \begin{subfigure}{2.3in}
      \centering
    \includegraphics[width=1.0\columnwidth]{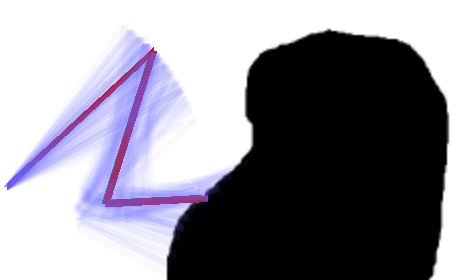}
  \end{subfigure} 
    \begin{subfigure}{2.3in}
      \centering
    \includegraphics[width=1.0\columnwidth]{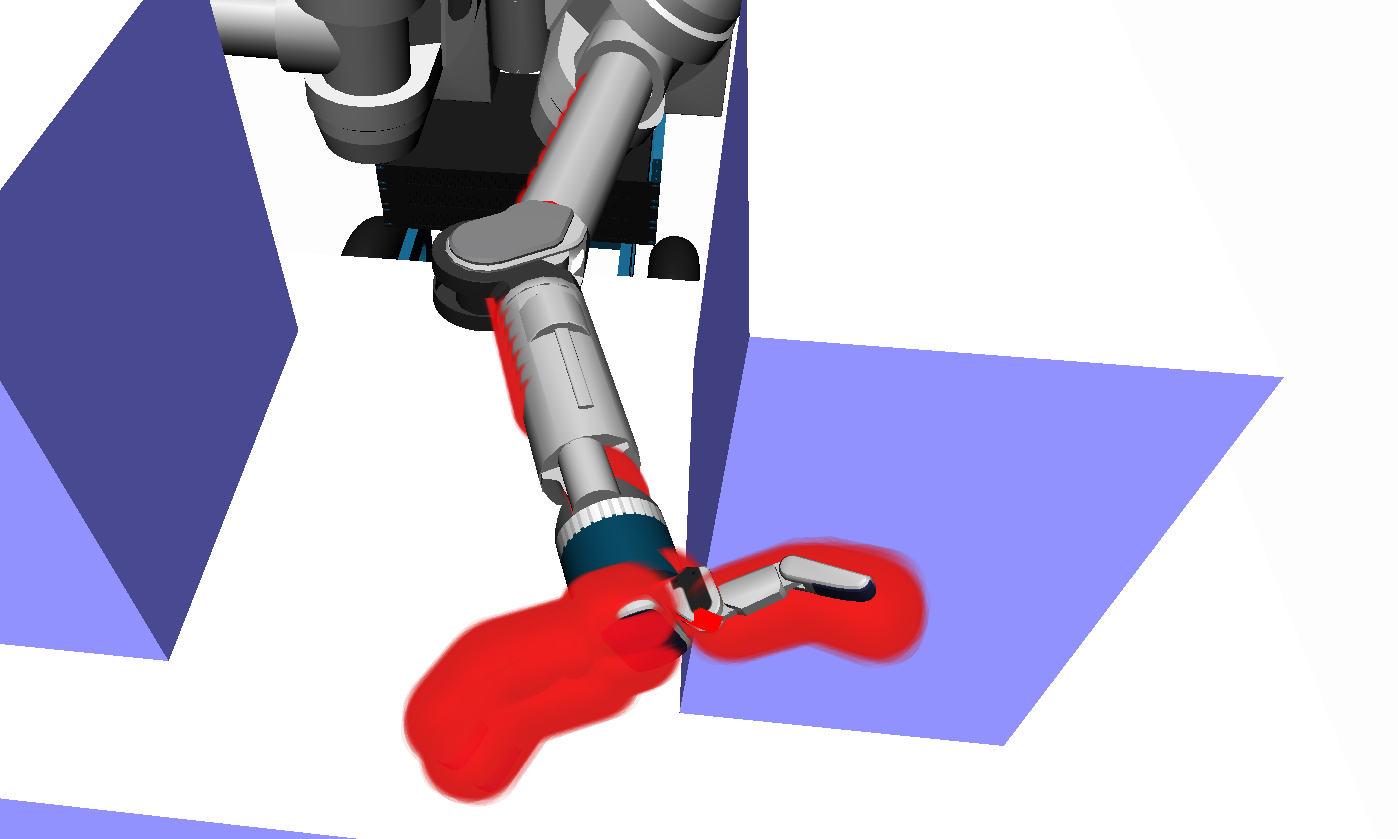}
  \end{subfigure} 
  \begin{subfigure}{2.3in} 
    \centering
    \includegraphics[width=1.0\textwidth]{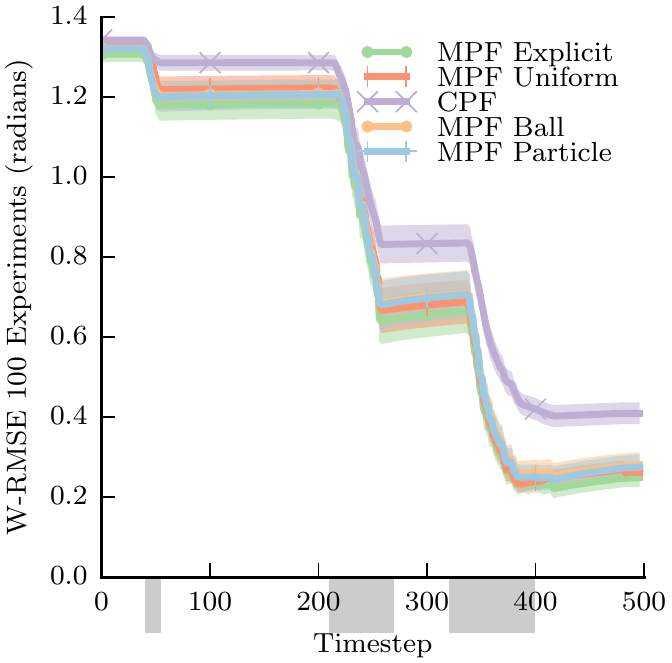}
    \caption{2-DOF Simulation.}
       \label{fig:simulation-2dof}
  \end{subfigure}
    \begin{subfigure}{2.3in}
      \centering
    \includegraphics[width=1.0\columnwidth]{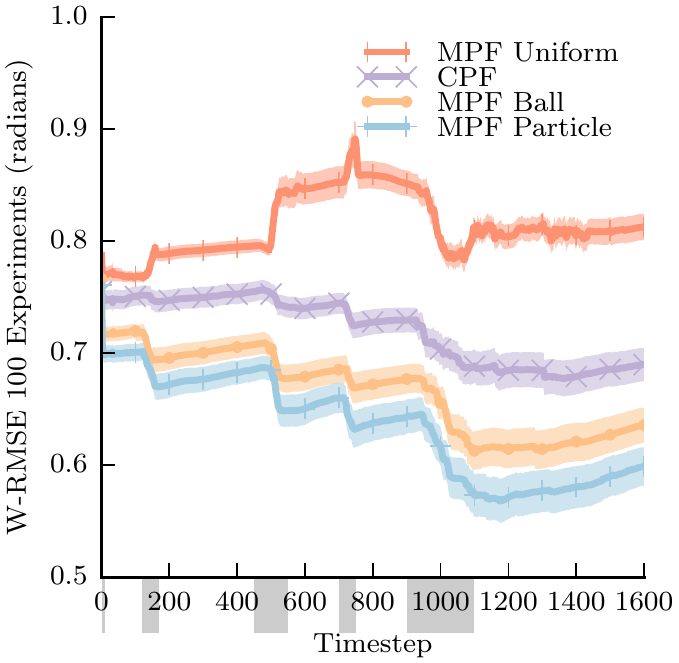} 
    \caption{3-DOF Simulation.}
      \label{fig:simulation-3dof} 
  \end{subfigure}
    \begin{subfigure}{2.3in}
      \centering
    \includegraphics[width=1.0\columnwidth]{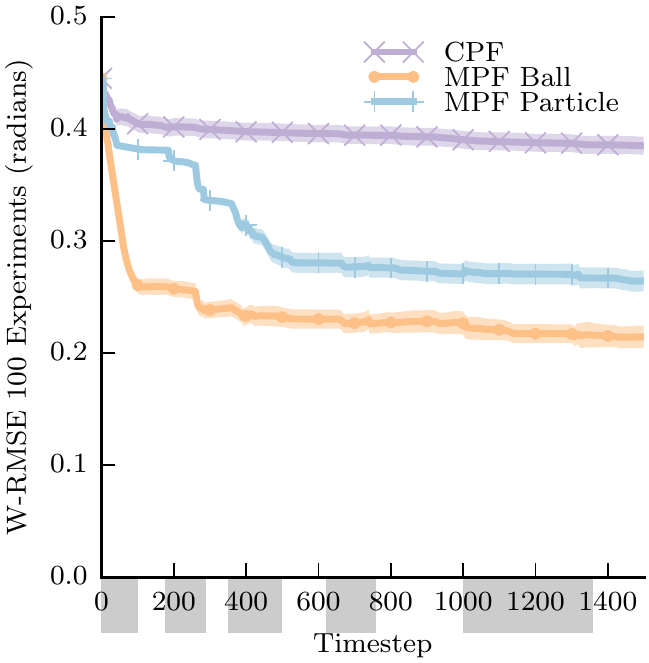}
    \caption{7-DOF Simulation.} 
     \label{fig:simulation-7dof} 
  \end{subfigure} 
  \caption{Filter performance using simulated data with 250 particles. Top row: the 2-DOF, 3-DOF and 7-DOF simulations. Bottom row: filter performance. Periods of persistent contact (in any experiemnt) are shaded grey on the $x$ axis. The weighted root mean square error is shown over 100 experiments with randomized starting conditions drawn from the prior, the $95 \%$ confidence interval is lightly shaded.}
\end{figure*}

\section{Simulation Experiments} 
\label{sec:experiments}

\subsection{Experimental Design}
We evaluate an estimator by performing a number of \emph{trials} where we draw
an initial state  $\state_0 \sim \bel(\state_0)$ from the initial belief state
$\bel(\state_0)$ and forward-simulate $\state_{t+1} \sim p(\state_{t+1} |
\state_t, \action_t)$ through a pre-defined sequence of $T$ actions $\action_1,
\dotsc, \action_T$. After each timestep, we draw an observation $\obs_t \sim
p(\obs_t | \state_t, \action_t)$ from the observation model.

The estimator returns a set of particles $\particles_t$ at each timestep. We
measure the accuracy of this estimate by computing the \emph{weighted root mean
square error} (W-RMSE)
\begin{align*}
  \text{W-RMSE}(\state_t, \particles_t) &= \sqrt{
    \frac{
      \sum_{i = 1}^k w_t^{[i]} ||\state_t^{[i]} - \state_t||^2
    }{
      \sum_{i = 1}^k w_t^{[i]}
    }
  }.
\end{align*}
We average W-RMSE over 100~trials with different initial states.

We considered three different environments:

\subsection{Two-Dimensional 2-DOF Arm}
\label{sec:experiment-2dof}
First, we consider a two degree-of-freedom arm in a two-dimensional environment
containing a single point obstacle (\figref{fig:simulation-2dof}). We set $\Sigma_\offset = 2.0 (\mathbf{I})$.
The robot executes a series of velocity commands that bring it into contact with the obstacle several times. 
The noise in the motion model $r_a = 0.05$~rad. It tracks its belief with $k = 250$ particles.

The robot has a point contact sensor on the tip of its manipulator. The contact
manifold always consists of two points in configuration space. Our simulation
results confirm this: the robot was able to reduce most of its uncertainty in
two touches. The first touch constrains the robot to one of two configurations.
The second touch disambiguates between those configurations. This also allows
us to implement MPF-Explicit by solving for an analytic solution to the robot's
inverse kinematics.

The CPF performed poorly due to particle deprivation. The contact
manifold has zero measure, so the probability of sampling a particle near the
manifold is low and the CPF collapses to one (possibly incorrect) mode. All MPF
variants significantly outperform the CPF, but perform similarly: the contact
manifold is small, so a small number of samples is sufficient to densely cover
it.

\subsection{Two-Dimensional 3-DOF Arm}
\label{sec:experiment-3dof}
Next, we consider a three degree-of-freedom arm in a two-dimensional
environment that contains a large, unstructured obstacle
(\figref{fig:simulation-3dof}). The robot has 20 circular contact sensors
spaced evenly along its outer two links. We set $\Sigma_\offset = 0.8 (\mathbf{I})$
The robot executes a series of human-controlled velocity commands that cause it to
come into contact with and, in some cases,slide along the obstacle. 
The motion model noise $r_a = 0.05$~rad. The robot tracks its belief with $k = 250$ particles.

The results show that MPF-Particle and MPF-Ball perform significantly better than
CPF, but MPF-Uniform performs worse. We cannot implement MPF-Explicit in this
domain because of the complex shape of the obstacle. MPF-Uniform suffers from
particle deprivation because the contact manifold is too large to cover densely
with $250$ particles.

Surprisingly, MPF-Particle slightly outperforms MPF-Ball in this domain. We
hypothesize that occurs because the robot maintains long periods of persistent
contact with the obstacle. Prior work has shown that the kernel density
estimation step of the MPF introduces additional variance into the belief state
during persistent contact~\cite{koval2015manifold_ijrr}. This is partially
masked by the local optimization performed by MPF-Particle.

\subsection{Three-Dimensional 7-DOF Arm}
\label{sec:experiment-7dof}
Finally, we consider a simulated seven degree-of-freedom Barrett
WAM~\cite{salisbury1988preliminary} equipped with a
BarrettHand~\cite{townsend2000barretthand} end-effector simulated by the
Dynamic Animation and Robotics Toolkit (DART)~\cite{unknown2015dynamic}. The
BarrettHand is kept in a fixed configuration and the environment contains of
two large boxes in front of the robot.  The robot has spherical contact sensors
(not shown in the figure) on its fingers, wrist, and forearm.  We set $\Sigma_\offset = 0.5(\mathbf{I})$.
The robot executes a deterministic series of velocity commands that cause it to touch the 
environment and slide along it. A small amount of uniform noise ($r_a = 0.01$~rad) is used
in the motion model. 250 particles are used to track the belief.

The results show that MPF-Ball and MPF-Particle both outperform CPF. It is not possible
to implement MPF-Explicit in this domain.  MPF-Uniform is omitted from 
the plot because its large error would distort the scale. In
this domain, MPF-Ball significantly outperforms MPF-Particle because the bias
introduced by a direct projection step prevents the filter from finding a good
solution. All filters perform in real-time (\tableref{table:timing}), with most of the
time spent computing the observation model (resampling and projection to the contact manifold). 
MPF-Uniform is particularly slow because it must continually reject samples that fail to project to the manifold.

\begin{table}
	\centering
	\begin{tabular} {l r r r}
	\hline
	Algorithm    & Total          & Transition   & Observation \\
	\hline 
    CPF          & $4 \pm 1$ ms   & $3 \pm 1$ ms & $1 \pm 1$ ms \\
	MPF-Particle & $9 \pm 7$ ms   & $2 \pm 1$ ms & $8 \pm 7$ ms \\
	MPF-Ball     & $10 \pm 7$ ms  & $2 \pm 1$ ms & $8 \pm 8$ ms \\
	MPF-Uniform  & $35 \pm 12$ ms & $2 \pm 1$ ms & $34 \pm 12$ ms \\
	\end{tabular}
	\caption{Timing data for the 7-DOF simulation (\figref{fig:simulation-7dof}), times shown in milliseconds with $95\%$ confidence interval. The total time is broken into its component transition and observation model updates. Only iterations of the filter where the robot was in contact are recorded. Time was recorded on a consumer laptop with an Intel core i7 CPU.}
	\label{table:timing}
\end{table}

\section{Real-Robot Experiments}
\label{sec:robot_experiments}
We validated our simulation results in two real-robot experiments on a 7-DOF
Barrett WAM~\cite{salisbury1988preliminary} equipped with a
BarrettHand~\cite{townsend2000barretthand} end-effector; the same platform
simulated in the 7-DOF simulation experiments. Each finger of the BarrettHand's
three fingers contain a strain gauge that measures torque around the distal
joint. The robot detects contact with the environment by thresholding changes
in measurements reported by these sensors. We treat this value as a binary
observation of contact anywhere on the distal finger link.

\subsection{Arm Configuration Estimation}
In the first experiment, we consider the robot's arm configuration to be
uncertain. We tele-operated the robot to execute a trajectory in an environment
similar to the one described in \sref{sec:experiment-7dof}. We use the
positions measured by the WAM's motor encoders, which are located before the
cable drive transmission, as proprioceptive observations $\obsq$. We measure
the ground truth position of the arm using optical joint encoders installed
after the cable drive transmission. These measurements are only used for
evaluation purposes and are not available to the estimator.

\begin{figure}
\centering
\begin{subfigure}{0.49\columnwidth}
  \centering 
  \includegraphics[width=1.0\columnwidth]{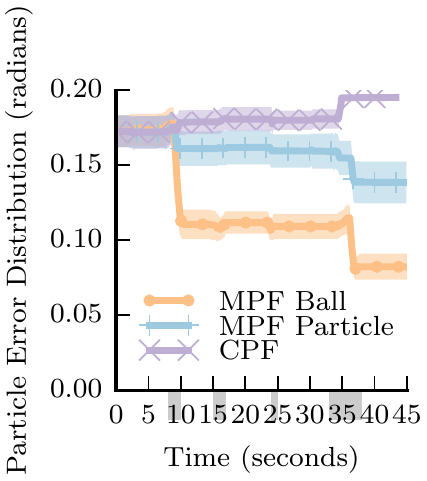}
  \caption{Noisy arm.}
  \label{fig:noisy_arm_experiment}
\end{subfigure}
\begin{subfigure}{0.49\columnwidth}
  \centering
  \includegraphics[width=1.0\columnwidth]{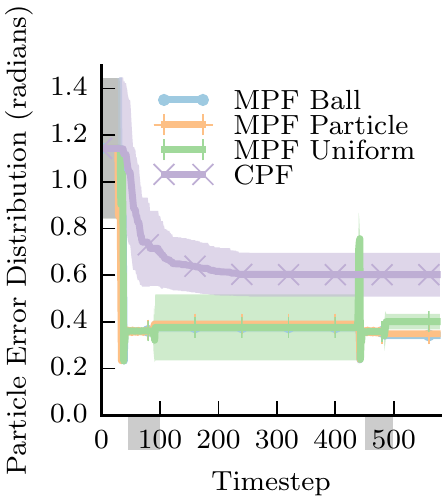}
  \caption{Underactuated hand.}
  \label{fig:underactuated_hand}
\end{subfigure}
\caption{Experiments with real robot data (\figref{fig:robot_experiments}) with 250 particles. 
One run is shown with the mean error and standard deviation of the particle distribution for each filter. 
The $x$-axis is lightly shaded in grey wherever the robot is in persistent contact.}
\end{figure}

Our goal is to estimate the 7-DOF configuration of the arm. Here, $\Sigma_\offset = 0.1 (\mathbf{I})$.
\figref{fig:noisy_arm_experiment} shows one run of the experiment with $k =
100$ particles, with $r_a = 0.01$ ~rad.  The results from this single experiment 
mirror those from \sref{sec:experiment-7dof}: both variants of 
MPF outperform CPF and MPF-Ball significantly outperforms MPF-Particle.

\subsection{Underactuated Hand State Estimation} 
\label{sec:underactuated_hand_experiment}
In the second experiment, the arm is held in a static configuration and the
robot closes its hand around an object with a known pose. The
BarrettHand~\cite{townsend2000barretthand} is underacutated: each finger has
two joints that are coupled by a mechanical clutch. We assume that the proximal
joint angles are known, but the distal joint angles are not. We record the
ground truth distal joint angles using joint encoders for evaluation purposes,
but do not make this data available to the estimator. The resolution of the 
SDF voxel grid was set to $10$~mm.

Our goal is to estimate the 3-DOF configuration of the distal finger joints. We
chose the initial belief state to be a uniform distribution over a ball with
radius $1$~rad centered at the origin of configuration space.
\figref{fig:underactuated_hand} shows data from one trial that contains two
grasps: one from $t \approx 40$ to $t \approx 100$ and another from  $t \approx
450$ to $t \approx 500$.

The data shows that all variants of the MPF significantly outperform the CPF,
but there is no significant difference between MPF-Uniform, MPF-Particle, and
MPF-Ball. This is consistent with our results in the two-dimensional planar
domain (\sref{sec:experiment-2dof}): this is a relatively simple problem
because each of the three joints is part of an independent kinematic chain.

We were surprised to see that the touches introduced large transients in the
MPF particle distribution, especially for MPF-Uniform. We believe that these
are caused by latency in the strain gauge sensors that is not accounted for in our
observation model.

\section{Discussion and Future Work}
\label{sec:limitations}

We have shown how the MPF can be extended to high dimensional state spaces by
implicitly representing the contact manifold with an objective function. Our
simulation and real-robot results show that this approach outperforms the 
conventional particle filter in a number of scenarios. This approach can
be used to compensate for proprioceptive error in a manipulator or to estimate
the configuration of an underactuated robotic hand.

However, our approach has a key limitation: it requires a known, static
environment. We could relax this requirement by incorporating the pose of
movable objects in the environment---including the base pose of the
robot---into the estimator's state space. This is challenging because: (1) it
is no longer possible to pre-compute a signed distance field and (2) the
behavior of the projection depends on the parameterization of this
configuration state space. We plan to explore these challenges in future work.

Finally, even though an implicit representation of the contact manifold allows
us to sample from it efficiently, the samples we draw are biased by the fact
that uniform samples of the ambient space will not project uniformly onto the
contact manifold for two reasons: (1) because the measures of the full state space
and the contact manifold may differ (a problem that even an explicit representation
of the contact manifold suffers from) and (2) because projecting samples from the full
state space to the contact manifold introduces bias. Our experiments have so far not made
clear to what extent this bias degrades performance of the filter. One way of eliminating
bias might be by rejecting and resampling particles that fail to meet some criterion
of uniformity (\eg Poisson disc sampling). An alternative solution might be sampling from the
tangent space of the manifold rather than projecting to it~\cite{Snoussi2013}.

\bibliographystyle{IEEEtran}
\bibliography{pr-refs.bib,mkoval.bib} 

\begin{thebibliography}{10}
\providecommand{\url}[1]{#1}
\csname url@samestyle\endcsname
\providecommand{\newblock}{\relax}
\providecommand{\bibinfo}[2]{#2}
\providecommand{\BIBentrySTDinterwordspacing}{\spaceskip=0pt\relax}
\providecommand{\BIBentryALTinterwordstretchfactor}{4}
\providecommand{\BIBentryALTinterwordspacing}{\spaceskip=\fontdimen2\font plus
\BIBentryALTinterwordstretchfactor\fontdimen3\font minus
  \fontdimen4\font\relax}
\providecommand{\BIBforeignlanguage}[2]{{%
\expandafter\ifx\csname l@#1\endcsname\relax
\typeout{** WARNING: IEEEtran.bst: No hyphenation pattern has been}%
\typeout{** loaded for the language `#1'. Using the pattern for}%
\typeout{** the default language instead.}%
\else
\language=\csname l@#1\endcsname
\fi
#2}}
\providecommand{\BIBdecl}{\relax}
\BIBdecl

\bibitem{koval2015manifold_ijrr}
M.~Koval, N.~Pollard, and S.~Srinivasa, ``Pose estimation for planar contact
  manipulation with manifold particle filters,'' \emph{International Journal of
  Robotics Research}, vol.~34, no.~7, pp. 922--945, 2015.

\bibitem{salisbury1988preliminary}
K.~Salisbury, W.~Townsend, B.~Eberman, and D.~DiPietro, ``Preliminary design of
  a whole-arm manipulation system ({WAMS}),'' in \emph{{IEEE} International
  Conference on Robotics and Automation}, 1988.

\bibitem{townsend2000barretthand}
W.~Townsend, ``The {BarrettHand} grasper--programmably flexible part handling
  and assembly,'' \emph{Industrial Robot: An International Journal}, vol.~27,
  no.~3, pp. 181--188, 2000.

\bibitem{boots2014learning}
B.~Boots, A.~Byravan, and D.~Fox, ``Learning predictive models of a depth
  camera \& manipulator from raw execution traces,'' in \emph{{IEEE}
  International Conference on Robotics and Automation}, 2014.

\bibitem{simunovic79information}
S.~Simunovic, ``An information approach to parts mating,'' Ph.D. dissertation,
  Massachusetts Institute of Technology, 1979.

\bibitem{kalman1960kf}
R.~Kalman, ``A new approach to linear filtering and prediction problems,''
  \emph{Journal of Basic Engineering}, 1960.

\bibitem{julier1997ukf}
S.~Julier and J.~Uhlmann, ``A new extension of the {K}alman filter to nonlinear
  systems,'' in \emph{International Symposium on Aerospace/Defense Sensing,
  Simulation, and Controls}, 1997.

\bibitem{gordon1993novel}
N.~Gordon, S.~D.J., and A.~Smith, ``Novel approach to nonlinear/non-{G}aussian
  {B}ayesian state estimation,'' in \emph{{IEE} Proceedings {F}}, 1993.

\bibitem{petrovskaya2011global}
A.~Petrovskaya and O.~Khatib, ``Global localization of objects via touch,''
  \emph{{IEEE} Transactions on Robotics}, vol.~27, no.~3, pp. 569--585, 2011.

\bibitem{javdani2013efficient}
S.~Javdani, M.~Klingensmith, J.~Bagnell, N.~Pollard, and S.~Srinivasa,
  ``Efficient touch based localization through submodularity,'' in \emph{{IEEE}
  International Conference on Robotics and Automation}, 2013.

\bibitem{hebert2013next}
P.~Hebert, T.~Howard, N.~Hudson, J.~Ma, and J.~Burdick, ``The next best touch
  for model-based localization,'' in \emph{{IEEE} International Conference on
  Robotics and Automation}, 2013.

\bibitem{smallwood1973optimal}
R.~Smallwood and E.~Sondik, ``The optimal control of partially observable
  {M}arkov processes over a finite horizon,'' \emph{Operations Research},
  vol.~21, no.~5, pp. 1071--1088, 1973.

\bibitem{hsiao2007grasping}
K.~Hsiao, L.~Kaelbling, and T.~Lozano-P\`{e}rez, ``Grasping {POMDP}s,'' in
  \emph{{IEEE} International Conference on Robotics and Automation}, 2007.

\bibitem{horowitz2013interactive}
M.~Horowitz and J.~Burdick, ``Interactive non-prehensile manipulation for
  grasping via {POMDP}s,'' in \emph{{IEEE} International Conference on Robotics
  and Automation}, 2013.

\bibitem{koval2014precontact_rss}
M.~Koval, N.~Pollard, and S.~Srinivasa, ``Pre- and post-contact policy
  decomposition for planar contact manipulation under uncertainty,'' in
  \emph{Robotics: Science and Systems}, 2014.

\bibitem{zhang2012application}
L.~Zhang and J.~Trinkle, ``The application of particle filtering to grasping
  acquisition with visual occlusion and tactile sensing,'' in \emph{{IEEE}
  International Conference on Robotics and Automation}, 2012.

\bibitem{zhang2013dynamic}
L.~Zhang, S.~Lyu, and J.~Trinkle, ``A dynamic {B}ayesian approach to
  simultaneous estimation and filtering in grasp acquisition,'' in \emph{{IEEE}
  International Conference on Robotics and Automation}, 2013.

\bibitem{pellegrini2008generalisation}
S.~Pellegrini, K.~Schindler, and D.~Nardi, ``A generalisation of the {ICP}
  algorithm for articulated bodies,'' in \emph{British Machine Vision
  Conference}, vol.~3, 2008, p.~4.

\bibitem{krainin2011manipulator}
M.~Krainin, P.~Henry, X.~Ren, and D.~Fox, ``Manipulator and object tracking for
  in-hand {3D} object modeling,'' \emph{International Journal of Robotics
  Research}, vol.~30, no.~11, pp. 1311--1327, 2011.

\bibitem{klingensmith2013closed}
M.~Klingensmith, T.~Galluzzo, C.~Dellin, M.~Kazemi, J.~Bagnell, and N.~Pollard,
  ``Closed-loop servoing using real-time markerless arm tracking,'' in
  \emph{{IEEE} International Conference on Robotics and Automation Humanoids
  Workshop}, 2013.

\bibitem{schmidt2014dart}
T.~Schmidt, R.~Newcombe, and D.~Fox, ``{DART}: Dense articulated real-time
  tracking,'' in \emph{Robotics: Science and Systems}, 2014.

\bibitem{schmidt2015depth}
T.~Schmidt, K.~Hertkorn, R.~Newcombe, Z.~Marton, M.~Suppa, and D.~Fox,
  ``Depth-based tracking with physical constraints for robot manipulation,'' in
  \emph{{IEEE} International Conference on Robotics and Automation}, 2015.

\bibitem{dogar2010proprioceptive}
M.~R. Dogar, V.~Hemrajani, D.~Leeds, B.~Kane, and S.~Srinivasa,
  ``Proprioceptive localization for mobile manipulators,'' Robotics Institute,
  Carnegie Mellon University, Tech. Rep. {CMU-RI-TR-10-05}, 2010.

\bibitem{chitta2007proprioceptive}
P.~V. Sachin Chitta~and, R.~Geykhman, and D.~D. Lee, ``Proprioceptive
  localization for a quadrupedal robot on known terrain,'' in \emph{{IEEE}
  International Conference on Robotics and Automation}, 2007.

\bibitem{kry2006interaction}
P.~G. Kry and D.~K. Pai, ``Interaction capture and synthesis,'' \emph{{ACM}
  Transactions on Graphics}, vol.~25, no.~3, 2006.

\bibitem{chung2015quadratic}
S.-J. Chung, J.~Kim, S.~Han, and N.~S. Pollard, ``Quadratic encoding for hand
  pose reconstruction from multi-touch input,'' in \emph{Eurographics Short
  Paper}, 2015.

\bibitem{ha2011human}
S.~Ha, Y.~Bai, and C.~K. Liu, ``Human motion reconstruction from force
  sensors,'' in \emph{{ACM SIGGRAPH}/Eurographics Symposium on Computer
  Animation}, 2011.

\bibitem{roncone2014automatic}
A.~Roncone, M.~Hoffmann, U.~Pattacini, and G.~Metta, ``Automatic kinematic
  chain calibration using artificial skin: self-touch in the icub humanoid
  robot,'' in \emph{Robotics and Automation (ICRA), 2014 IEEE International
  Conference on}.\hskip 1em plus 0.5em minus 0.4em\relax IEEE, 2014, pp.
  2305--2312.

\bibitem{rosenblatt1956remarks}
M.~Rosenblatt, ``Remarks on some nonparametric estimates of a density
  function,'' \emph{The Annals of Mathematical Statistics}, vol.~27, no.~3, pp.
  832--837, 1956.

\bibitem{Tian2016116}
\BIBentryALTinterwordspacing
H.~Tian, X.~Zhang, C.~Wang, J.~Pan, and D.~Manocha, ``Efficient global
  penetration depth computation for articulated models,'' \emph{Computer-Aided
  Design}, vol.~70, pp. 116 -- 125, 2016, \{SPM\} 2015. [Online]. Available:
  \url{http://www.sciencedirect.com/science/article/pii/S0010448515001074}
\BIBentrySTDinterwordspacing

\bibitem{Felzenszwalb04}
P.~F. Felzenszwalb and D.~P. Huttenlocher, ``Distance transforms of sampled
  functions,'' Cornell Computing and Information Science, Tech. Rep., 2004.

\bibitem{ratliff2009chomp}
N.~Ratliff, M.~Zucker, J.~A. Bagnell, and S.~Srinivasa, ``Chomp: Gradient
  optimization techniques for efficient motion planning,'' in \emph{Robotics
  and Automation, 2009. ICRA '09. IEEE International Conference on}, May 2009,
  pp. 489--494.

\bibitem{unknown2015dynamic}
``{D}ynamic {A}nimation and {R}obotics {T}oolkit,''
  \url{http://dartsim.github.io}, 2013.

\bibitem{Snoussi2013}
H.~Snoussi, ``Particle filtering on riemannian manifolds. application to
  covariance matrices tracking,'' in \emph{Matrix Information Geometry}, 2013,
  pp. 427--449.

\end{thebibliography}
\end{document}